\documentclass[11pt,onecolumn]{article}
\usepackage{algorithm, algorithmic, setspace}
\usepackage{graphicx} 

\usepackage{amsmath} 
\usepackage{amssymb}  
\usepackage{amsthm}
\usepackage{amsfonts}
\usepackage{dsfont}
\usepackage{color}
\usepackage{subfig}
\usepackage[english]{babel}
\usepackage[latin1]{inputenc}
\newcommand{\fun}{\mathcal{F}}
\newcommand{\gun}{\mathcal{H}}

\newcommand{\soft}{\mathbb{S}}

\newcommand{\argmin}[1]{\underset{#1}{\mathrm{argmin\,}}}

\newcommand{\xtrue}{\widetilde{x}}

\newcommand{\R}{\mathds{R}}

\newtheorem{proposition}{Proposition}
\newtheorem{theorem}{Theorem}

\begin{document}
\title{A biconvex analysis for Lasso $\ell_1$ reweighting}
\author{Sophie M. Fosson}
\maketitle
\begin{abstract}Iterative $\ell_1$ reweighting algorithms are very popular in sparse signal recovery and compressed sensing, since in the practice they have been observed to outperform classical $\ell_1$ methods. Nevertheless, the theoretical analysis of their convergence is a critical point, and generally is limited to the convergence of the functional to a local minimum or to subsequence convergence. In this letter, we propose a new convergence analysis of a Lasso $\ell_1$ reweighting method, based on the observation that the algorithm is an alternated convex search for a biconvex problem. Based on that, we are able to prove the  numerical convergence of the sequence of the iterates generated by the algorithm. 
Furthermore, we propose an alternative iterative soft thresholding procedure, which is faster than the main algorithm.
\end{abstract}
\section{Introduction}
In the recent literature on compressed sensing (CS, \cite{don06}), much attention has been devoted to iterative $\ell_1$ reweighting (IRL1) techniques, see, \emph{e.g.},  \cite{can08rew,fou09,wip10,zhao12,lyu13,asi13,che14,ahm15}. In the practice, IRL1 algorithms are efficient and accurate; in theory, their convergence is guaranteed in the sense of local minimization of a certain non-convex functional, while the iterates are not proved to converge. This work goes one step further by proving the numerical convergence of the iterates, under mild conditions, leveraging a biconvex interpretation of the problem.

Hereinafter, we review the basics of CS in relation to $\ell_1$ minimization (Section \ref{sub:CS}), we introduce the IRL1 basics (Section \ref{sub:RE}), and we illustrate the related literature along with the problem of convergence (Section \ref{sub:REV}). Based on that, our contribution will be further specified in Section \ref{sub:CO}.
\subsection{Compressed sensing and $\ell_1$ minimization}\label{sub:CS}
CS states that, under certain conditions, a $k$-sparse signal $x\in\R^n$, \emph{i.e.}, a signal with only $k\ll n$ non-zero components, can be recovered from compressed, linear measurements $y=Ax$ (possibly corrupted by noise), where $A\in\R^{m,n}$, $m<n$. It is well known that CS can be re-formulated in a convex framework, using the $\ell_1$-norm to approximate $\ell_0$-norm, \emph{i.e.}, the number of non-zeros in a vector. In particular, the Basis Pursuit (BP) and the Basis Pursuit Denoising (BPDN) formulations have been introduced, which respectively read as follows: $\min_{x\in\R^n}\left\| x\right\|_1~~\text{s.t.}~~Ax=y$ and 
$\min_{x\in\R^n}\left\| x\right\|_1~~\text{s.t.}~~\left\|y-A x \right\|_2<\eta$, where $\eta>0$ is a parameter that takes account of measurement noise.
(see, \emph{e.g.}, \cite[Chapter 4]{fou11} for a thorough review). BP and BPDN can be solved using classic convex programming algorithms, \emph{e.g.}, the interior points method. A non-constrained version of BPDN, known as Lasso \cite{tib96}, is often considered: $\min_{x\in\R^n}\frac{1}{2}\left\|y-A x \right\|_2^2+\lambda \| x\|_1$, with $\lambda>0.$

Lasso was popular long before CS  as variable selection method: the $\ell_1$ regularization, in fact, even in overdetermined problems, can be used to find out the most relevant components in a vector. In CS, the Lasso formulation is desirable for different motivations, \emph{e.g.}, the possibility of solution through simple iterative methods, such as the iterative soft thresholding (IST, \cite{dau04, for10}) and alternating direction method of multipliers (ADMM, \cite{boy10}). IST and ADMM are easy to implement and are prone to parallel computation \cite{fia18} and distributed recovery, \emph{e.g.}, in sensor networks \cite{mot13,fox14,rfm15,mata15,fox16}.
\subsection{Enhancing $\ell_1$ minimization via reweighting}\label{sub:RE}
In the seminal work \cite{can08rew}, the question arose about how $\ell_1$-minimization could be improved, starting from the observation that $\ell_1$ is less "democratic" than $\ell_0$, that is, it penalizes more the coefficients with larger magnitude. This originated the attempt to reweight the $\ell_1$-norm term in order to improve the final estimation. The proposal in \cite{can08rew} was to use weights that are inversely proportional to the magnitudes of the corresponding coefficients. This was implemented through the following IRL1 algorithm for BP: given $x(0)\in\R^n$, for $t=1,\dots,T_{stop}$,
\begin{equation}\label{rwl1_can0}
\begin{split}
&w_i(t)= \frac{1}{ |x_i(t)|+\varepsilon} \text{ for any } i\in \{1,\dots, n\}\\
&x(t+1)= \argmin{x\in\R^n} \sum_{i=1}^n w_i(t) |x_i| \text{ s.t. }  y=A x
\end{split}
\end{equation}
where $\varepsilon>0$ is necessary to avoid infinite weights, and $T_{stop}$ is the time step at which a suitable convergence criterion is satisfied. This algorithm converges in the sense that given $g(|x_i|):=\log(|x_i|+\varepsilon)$, the functional $ \sum_{i=1}^n g(|x_i(t)|)$ converges to a (local) minimum of
\begin{equation}\label{concave}
\sum_{i=1}^n g(|x_i|)~~\text{ s.t. }~~ Ax=y.
\end{equation}
This was proved exploiting a Minimization-Maximization (MM, \cite{sun17}) perspective, as explained in \cite[Section 2.3]{can08rew}. Nevertheless, the iterates $x(t)$ are not guaranteed to converge. In the practice, this is a  drawback, because it requires the algorithm to compute $\sum_{i=1}^n g(|x_i(t)|)$ to evaluate $T_{stop}$.  If the iterates $x(t)$ were known to converge, one should simply check $|x_i(t)-x_i(t-1)|<\delta$ for a given tolerance $\delta$, singularly for each $i$, without computing the whole functional.
\subsection{Review of the literature and the convergence problem}\label{sub:REV}
Beyond \cite{can08rew}, reviewed in the previous section, other works have focused on iterative reweighting in the last decade. 

It is worth to mention that a popular class of reweighting methods is the so-called iteratively reweighted least squares (IRLS, \cite{irls}). IRLS techniques perform an $\ell_2$ reweighting, \emph{i.e.}, they adaptively modify the weights of a least squares functional, which has been proved to be efficient for sparse recovery \cite{fig07, cha08,rav15irls}. Their convergence is a tricky point: in general, they converge only to a set of points \cite{ba14}, while stronger convergence has been proved only in specific settings.

Coming back to $\ell_1$ reweighting, 
in \cite{nee09} the approach of \cite{can08rew} was analyzed in the noisy case. In \cite{fou09}, algorithm \eqref{rwl1_can0} was re-elaborated using weights $(|x_i(t)|+\epsilon(t))^{q-1}$, with $q\in(0,1)$ and non-increasing $\epsilon(t)$. \cite[Proposition 4.2]{fou09} proves the existence of a convergent subsequence. Convergence to the correct sparse signal was shown in numerical experiments.  The same method was analyzed in \cite{che14}, which provided some further characterization of the accumulation points.

In \cite{zhao12}, a unified theoretical analysis was proposed for a large class of IRL1 methods for noise-free, underdetermined linear systems. Convergence analyses were proposed for different concave \emph{merit} functions $F_{\epsilon}(x)=\sum_i g_i (|x_i|+\epsilon)$ \cite[Eq 2.1]{zhao12}, under some technical assumptions \cite[Assumption 2.1]{zhao12}. In general, convergence of the functional to a local minimum and subsequence convergence were achieved. For $F_{\epsilon}(x)$  bounded below for any $|x_i|>0$ and $\epsilon>0$, numerical convergence was also proved for an IRL1 procedure with decreasing $\epsilon$ \cite[Algorithm 2.2, Corollary 3.4]{zha12}. However, this is not valid in classical cases, \emph{e.g.}, for algorithm \eqref{rwl1_can0}. Moreover, vanishing parameters are generally not desired in all those applications where the signal is not finite in time and processing/recovery must be performed continuously.

In \cite{asi13}, homotopy strategies were elaborated for the problem $\min \sum_{i=1}^n w_i |x_i|+ \frac{1}{2}\|Ax-y\|_2^2 $, whose efficiency was verified via numerical experiments. Finally, in \cite{ahm15}, composite regularization was considered, and convergence was proved in the sense of asymptotic stationary point condition \cite[Definition 2.4]{mai15}, leveraging an MM interpretation. 

In conclusion, to the best of our knowledge, previous convergence analyses for IRL1 algorithms generally achieved the convergence of the functional to a stationary point or local minimum, while numerical convergence was shown only in a specific case with decreasing parameter $\epsilon$ \cite{zhao12}. We refer the interested reader to \cite{wip10,lyu13} for further comparisons between $\ell_1$ (IRL1) and $\ell_2$ (IRLS) reweighting methods.

{\color{black}{We finally notice that problem \eqref{concave} is concave. In the following, we will see that $\ell_1$ reweighting can tackle sparse optimization problems with concave penalization. The interest on sparse optimization with concave penalization has substantially increased in the last few years, and novel non-reweigthing techniques have been proposed \cite{par15, bayram16,fox16,sel17}. However, most of this literature \cite{par15,bayram16} does not envisage the compressed case (as it requires a non-singular $A^T A$). For reasons of space, we leave for future work an extended analysis of non-compressed case along with the comparison to novel non-reweighting approaches.}} 

\subsection{Contribution}\label{sub:CO}
In this letter, we propose a biconvex interpretation of a Lasso $\ell_1$ reweighting method. This new perspective allows us to prove the numerical convergence of the iterates, that is, $\lim_{t\to\infty}\left\|x(t)-x(t-1)\right\|_2=0$. This is not sequence convergence, \emph{i.e.}, we are not proving that $x(t)\to x^{\star}\in\R^n$. However, numerical convergence is sufficient for practical purposes, for example to evaluate $T_{stop}$.

The remainder of the letter is organized as follows. In Section \ref{sec:problem}, we state the main problem, the main algorithm, the biconvex interpretation, and our convergence results. Afterwards, in Section \ref{sec:simple}, we will propose a simplified algorithm which overcomes the problem of nested loops in the main algorithm and requires less iterations.
\section{Problem statement and theoretical analysis}\label{sec:problem}
As mentioned in \cite{can08rew}, problem \eqref{concave} can be formulated for any concave, non-decreasing $g(|x_i|)$ \cite{woo16}. In addition to $\log |x_i|+\varepsilon$, other popular $g$'s are $\ell_q$, with $q\in (0,1)$ \cite{cha08, irls, rav15irls, woo16}; smoothly clipped absolute deviations (SCAD) \cite{fan01_pioneer}; minimax concave penalty (MCP) \cite{zha10MCP, woo16, fox16, hua18}. Extending \cite{can08rew} to generic concave, non-decreasing $g$'s,  the following algorithm can be implemented to achieve a (local) minimum of
$  \sum_{i=1}^n g(| x_i|)$ s.t. $y=A x$: given $x(0)\in\R^n$, for any $t=1,\dots,T_{stop}$ 
\begin{equation}\label{rwl1_can}
\begin{split}
&w_i(t)=  g'(|x_i(t)|) \text{ for any } i\in \{1,\dots, n\}\\
&x(t+1)= \argmin{x\in\R^n} \sum_{i=1}^n w_i(t) |x_i|~~\text{ s.t. }~~  y=A x\\
\end{split}
\end{equation}
The analytical justification provided in \cite{can08rew} is as follows. Since $G=\sum_{i=1}^n g(| x_i|)$ is concave, thus below its tangent, one can improve on a guess $v\in\R^n$ by locally minimizing a linearization of $G$ around $v$. In this way, one obtains an MM procedure which corresponds to the algorithm \eqref{rwl1_can}. Leveraging this interpretation, the convergence of the algorithm can be proved using the Global Convergence Theorem \cite[Chapter 7]{lue16}, as also illustrated in \cite{faz03}. As remarked in \cite[Section 2.3]{can08rew}, convergence is here intended to a local minimum of the functional, while the iterates of the algorithm are not proved to converge.
Following this philosophy, let us reformulate the Lasso problem using a concave penalty (from now onwards, $\R_+=[0,+\infty)$):
\begin{equation}\label{concave_penalization}  
\begin{split}
 &\min_{x\in\R^n}\frac{1}{2}\left\|y-A x \right\|_2^2+\lambda \sum_{i=1}^n g(|x_i|),~~~\lambda>0\\
 &g:\R_+\to \R_+ \text{ concave, non-decreasing in } |x_i|.
\end{split}
\end{equation}
To tackle problem \eqref{concave_penalization}, we propose Algorithm \ref{rwl1_fox}, which is an analogous for Lasso of the algorithm \eqref{rwl1_can} for BP.
\begin{algorithm}
\setstretch{1.3}
  \caption{Lasso IRL1}\label{rwl1_fox}
  \begin{algorithmic}[1] 
  \STATE  Initialize: $x(0)\in\R^n$; $\lambda>0$
  \FORALL{$t=1,\dots,T_{stop}$}
  \STATE $w_i(t)=g'(|x_i(t)|)$ for any $i\in \{1,\dots, n\}$
  \STATE $x(t+1)= \argmin{x\in\R^n} \frac{1}{2}\left\|y-Ax\right\|_2^2 + \lambda\sum_{i=1}^n w_i(t) |x_i|$
  \ENDFOR     
  \end{algorithmic}
\end{algorithm} 
The convergence of Algorithm \ref{rwl1_fox} cannot be derived with the techniques used for algorithm \ref{rwl1_can}, as the functional \eqref{concave_penalization} is the sum of a convex and a concave terms, thus in general it is neither convex nor concave. The minimization on the local linearization then does not guarantee the descent. We now provide a convergence proof for Algorithm \ref{rwl1_fox}, based on biconvex minimization tools. 

\subsection{Biconvex perspective}\label{sub:bic}
Following the notation in \cite{gor07}, we remind that a function $f: X \times Y \to \R$ is \emph{biconvex} if it is convex in $y$ (respectively, in $x$) for any fixed $x\in X$ (respectively, $y\in Y$). The problem $$\min f(x,y)~~~(x,y)\in B,~B\subseteq X\times Y,~B \text{ biconvex }$$ is called a \emph{biconvex optimization problem} if $f$ is biconvex on $B$. We say that $(x^{\star}, y^{\star})\in B$ is a \emph{partial optimum} of $f$ on $B$ if $f(x^{\star}, y^{\star})\leq f(x, y^{\star})$, for  any $x\in B_{y^{\star}}$, and $f(x^{\star}, y^{\star})\leq f(x^{\star}, y)$, for  any $y\in B_{x^{\star}}$ where $B_{x}$ and $B_{y}$ respectively are the $x$-section and the $y$-section of $B$.

We name alternated convex search (ACS, \cite[Algorithm 4.1]{gor07}) the algorithm that iteratively minimizes $f(x,y)$ with respect to $x$ and $y$. Each iteration corresponds then to the solution of a convex problem, namely, $x(t+1)=\min_{x\in B_{y(t)}} f(x,y(t))$; $y(t+1)=\min_{y\in B_{x(t+1)}} f(x(t+1),y)$.

We can now prove our main results.
\begin{theorem}\label{theo:biconvexity}
Let $g(|x_i|)$ be strictly concave, increasing, and differentiable in its domain $D\subseteq \R_+$ (for $|x_i|=0$, the right derivative is considered).
Let us define the functional:
\begin{equation}\label{eq:biconvex}
\begin{split}
\fun&:X\times Y\to \R, ~~~X\subseteq R^n, Y\subseteq \R_+^n  \\
\fun(x,w)&:=\frac{1}{2}\left\|y-A x \right\|_2^2+\lambda\sum_{i=1}^n\left[w_i|x_i|+h(w_i)\right]
\end{split}
\end{equation}
where $h: H\subseteq \R_+\to \R_+$ is defined by $h'=-(g')^{-1}$,  $(\cdot)^{-1}$ indicating the inverse function. Then, $\fun$ is biconvex and Algorithm \ref{rwl1_fox} is an ACS for it.
\end{theorem}
Before proving the theorem, let us discuss the definitions of $h$, $X$, $Y$ for some popular $g$'s.

a) If $g(|x_i|)=\log(|x_i|+\epsilon)$ \cite{can08rew,cal16}, $g'(|x_i|)=(|x_i|+\epsilon)^{-1}$. Computing its inverse, we easily get $h'(w_i)=\epsilon-\frac{1}{w_i}$, from which $h(w_i)=\epsilon w_i-\log(w_i)$, modulus an integration constant which can be arbitrarily set (for instance, one can choose it so that $h(w_i)\geq 0$ to have a non-negative $\fun$). Since the corresponding $\fun$ is bounded below, continuous and coercive, given any compact set $S\subset \R^n\times \R^n$ of starting points, Algorithm \ref{rwl1_fox}  evolves in the compact set $\{(x,w): \fun(x,w)\leq \max_{s\in S} \fun(s)\}$. We can then assume that $X=[-\beta,\beta]^n$ for some sufficiently large $\beta>0$; thus, $Y=\left[(\beta+\epsilon)^{-1},{\epsilon}^{-1}\right]^n$. 

b) If $g(|x_i|)=(|x_i|+\epsilon)^q$, $q\in(0,1)$ \cite{fou09}, we obtain $h(w_i)=\epsilon w_i + \frac{1-q}{(w_i/q)^{q/(1-q)}}$. Again, we can set $X=[-\beta,\beta]^n$, $Y=\left[q/(\beta+\epsilon)^{1-q},q/\epsilon^{1-q}\right]^n$.

c) If $g(|x_i|)=\alpha|x_i|-\frac{1}{2}|x_i|^2$ with $\alpha>0$ and $X=\left[ -\alpha,\alpha\right]^n$  \cite{zha10MCP,fox16}, we can obtain $h(w_i)=\frac{1}{2}(\alpha-w_i)^2$, $Y=[0,\alpha]^n$.

\proof First of all, since $g$ is strictly concave, the derivative $g'$ is strictly monotonically decreasing. Therefore $g'$ is invertible and $h'$ is well defined. Moreover, $h'$ turns out to be strictly monotonically increasing, thus $h$ is strictly convex. As a consequence, $\fun$ is convex in $w$. As the convexity in $x$ is evident, we conclude that $\fun$ is biconvex.

Regarding the ACS of $\fun$, we notice that the minimization with respect to $x$ simply is a (weighted) Lasso, as in the first step of Algorithm \ref{rwl1_fox}.
To minimize with respect to $w$, we derive with respect to $w_i$, $i\in\{1,\dots,n\}$, and observe that the gradient is equal to zero whenever $|x_i|=-h'(w_i)$ for any $i$. By inversion we obtain $w_i=g'(|x_i|)$, which corresponds to the second step of Algorithm \ref{rwl1_fox}. This proves the thesis.\qed 

{\color{black}{We specify  that \eqref{eq:biconvex} is not derived from  \eqref{concave_penalization}, and the two functionals are not said to be equivalent. The properties of \eqref{eq:biconvex} as cost functional are not of our interest: we instead exploit \eqref{eq:biconvex} to study the convergence properties of Algorithm \ref{rwl1_fox}.}}

The fact that Algorithm \ref{rwl1_fox} is an ACS for a biconvex functional yields to the following theorem. 

\begin{theorem}\label{theo:numericalconvergence}
Let $\fun$ be as defined in \eqref{eq:biconvex}. Let us assume that  $X$ and $Y$ are closed, and that Lasso in Algorithm \ref{rwl1_fox} has a unique solution. Then, Algorithm \ref{rwl1_fox} achieves a partial optimum of $\fun$. Moreover, the iterates of Algorithm \ref{rwl1_fox} numerically converge, that is, $\lim_{t\to \infty}\left\|(x(t+1),w(t+1))-(x(t),w(t))\right\|_2=0$.
\end{theorem}
Theorem \ref{theo:numericalconvergence} is a consequence of the following result.
\begin{theorem}\label{theo:4.9} (\cite[Theorem 4.9]{gor07})
Let $f:X\times Y\to \R$ be biconvex, continuous, and let $X$ and $Y$ be closed. If the sequence generated by ACS is contained in a compact set, then it has a least one accumulation point $(x^{\star}, y^{\star})$. Moreover, if the convex problems  $\min_{x} f(x,y^{\star})$ and $\min_{y} f(x^{\star},y)$ both admit unique solutions, then all the accumulation points are partial optima and they form a connected compact set, and the sequence is numerically convergent.
\end{theorem}
Concerning the closedness of $X$ and $Y$, we have previously discussed in the examples a)-b)-c) the fact that for popular $g$'s we can work on compact sets. Moreover, uniqueness for the Lasso step is  guaranteed for most used sensing matrices, \emph{e.g.}, random sensing matrices generated from continuous distributions (see \cite{tib13}). On the other hand, $h(w_i)$ is strictly convex, hence $\fun$ has a unique minimum with respect to each $w_i$. 

In conclusion, Theorem \ref{theo:numericalconvergence} is valid for a wide class of popular IRL1 methods.

\section{A simple IST variant}\label{sec:simple}
\begin{figure}[ht]
\centering
\subfloat[][Noise-free]{\includegraphics[width=0.7\textwidth]{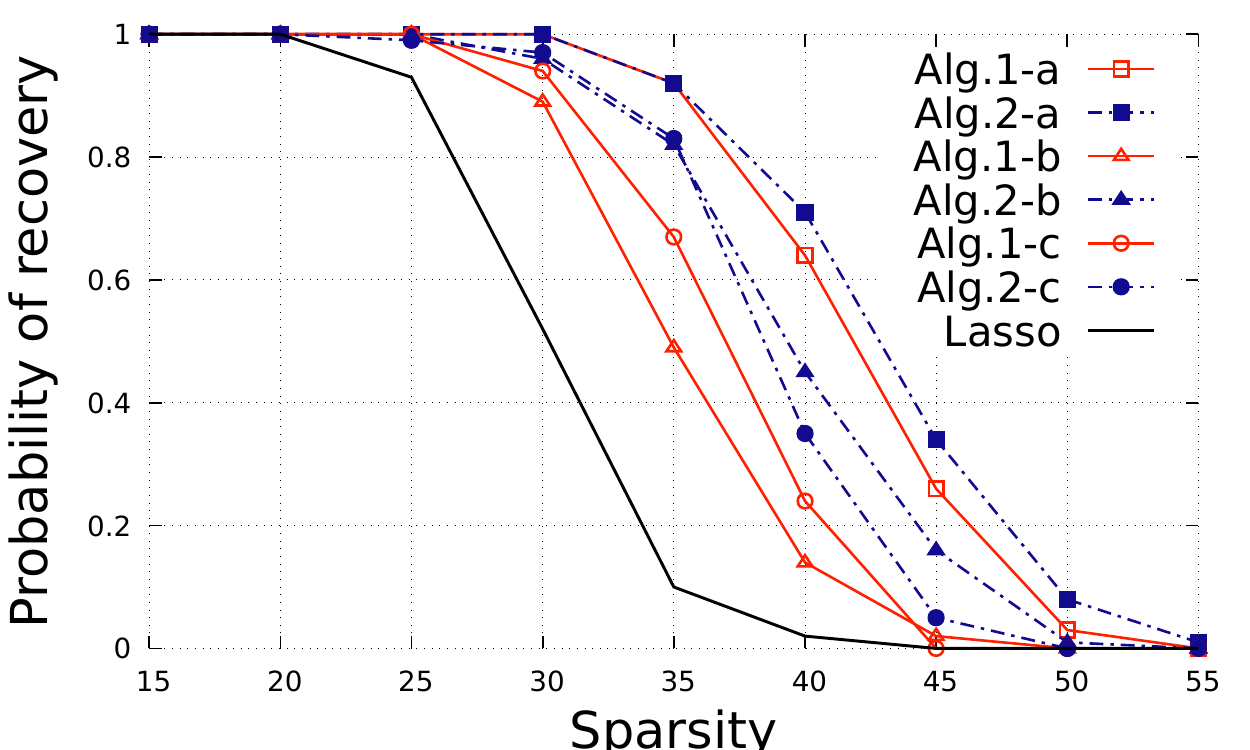}}$~~$
\subfloat[][Noise-free]{\includegraphics[width=0.7\textwidth]{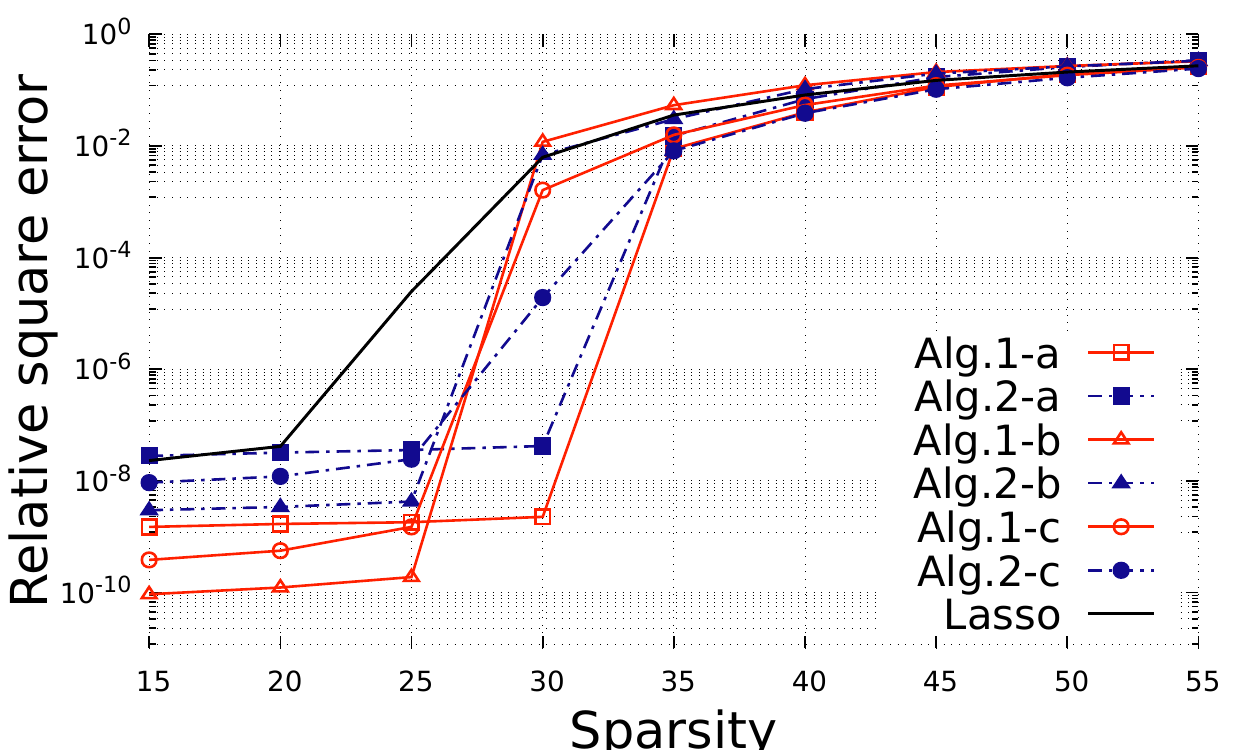}}$~~$
\subfloat[][Noise-free]{\includegraphics[width=0.7\textwidth]{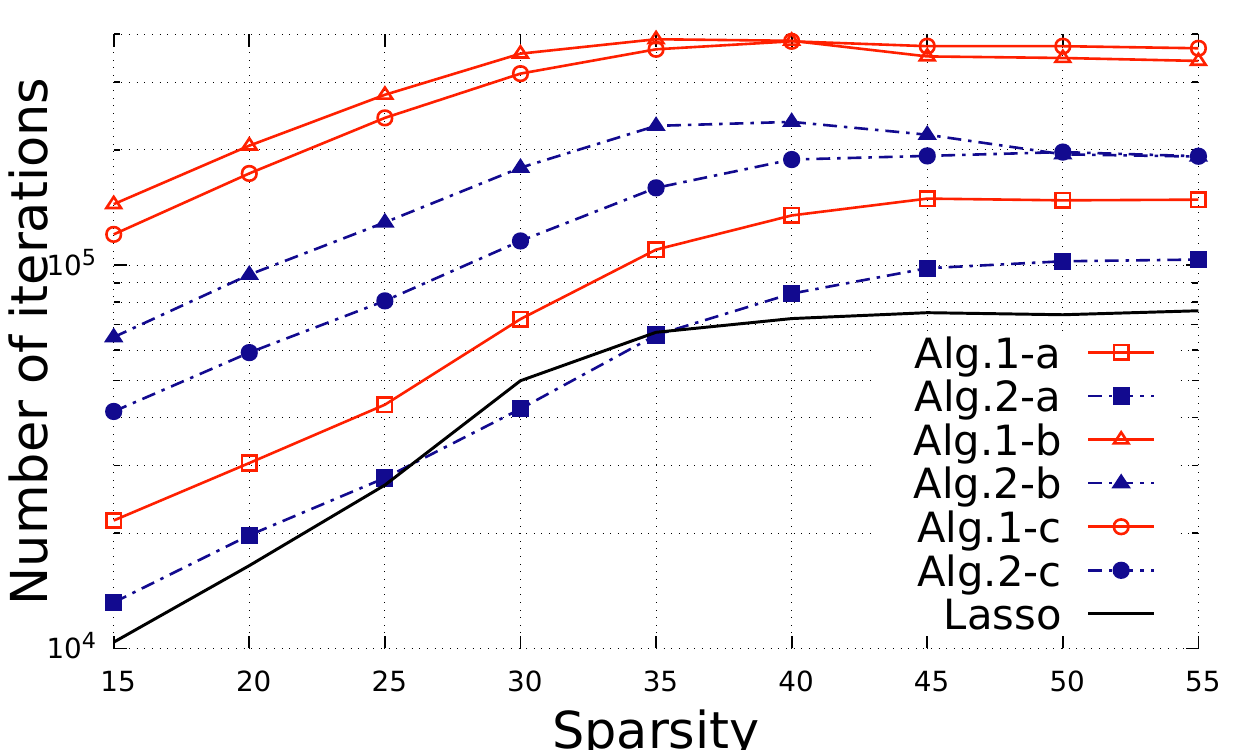}}\\
\caption{Lasso (via ADMM) vs Algorithms \ref{rwl1_fox} and \ref{rwl1_foxist}; $n=256$, $m=100$.}\label{fig:1}
\end{figure}
\begin{figure}[ht]
\centering
\subfloat[][SNR=25 dB]{\includegraphics[width=0.7\textwidth]{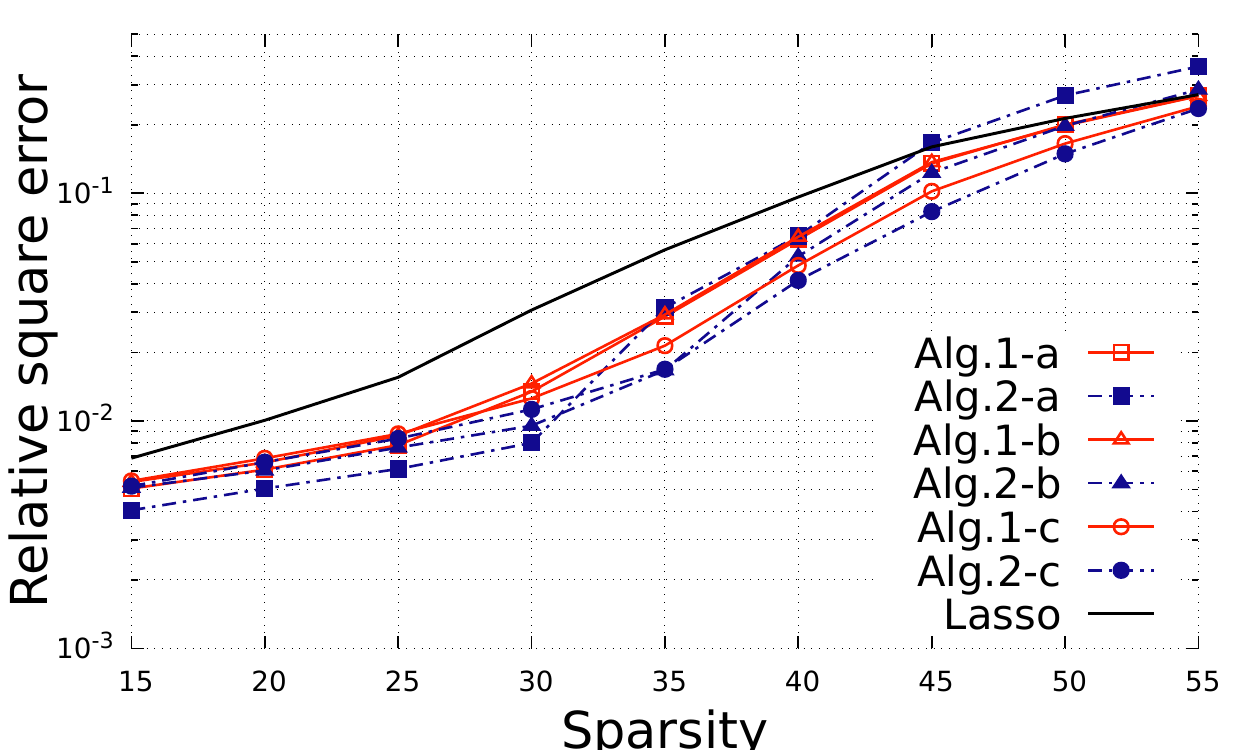}}$~~$
\subfloat[][SNR=25 dB]{\includegraphics[width=0.7\textwidth]{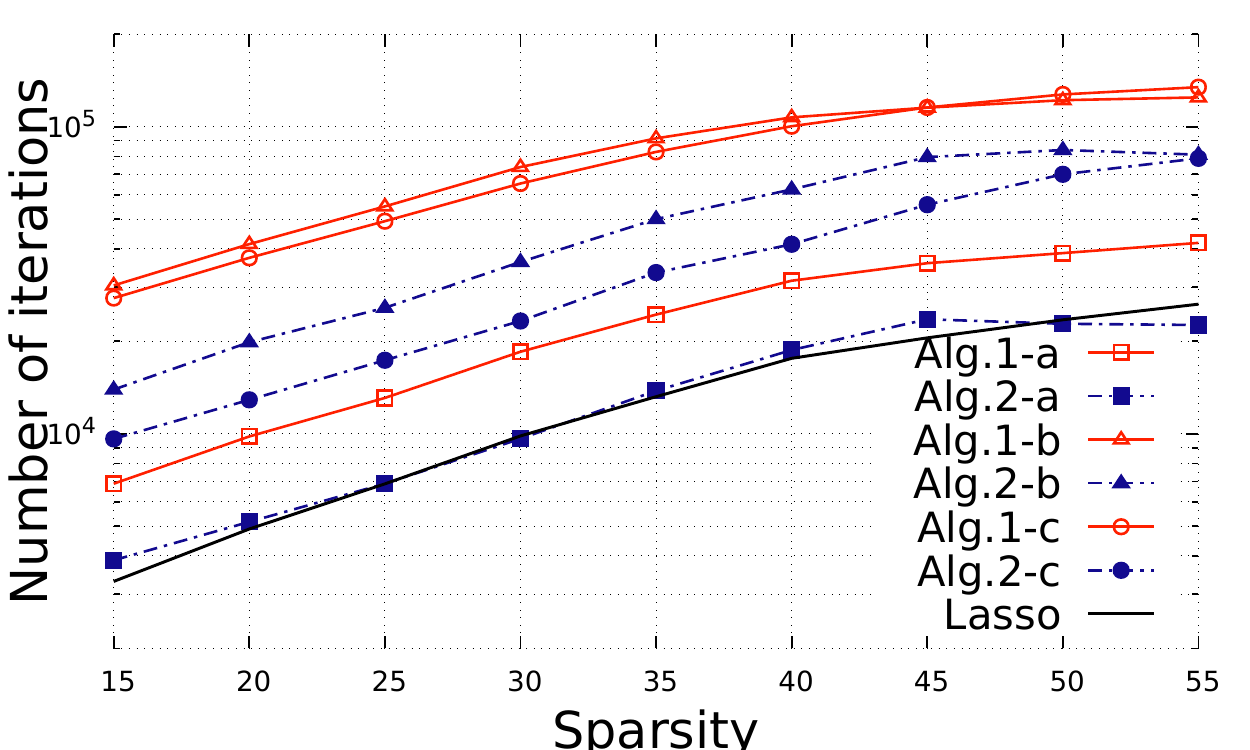}}$~~$
\subfloat[][Ranges for run times ('n' is for noisy setting)]{\includegraphics[width=0.7\textwidth]{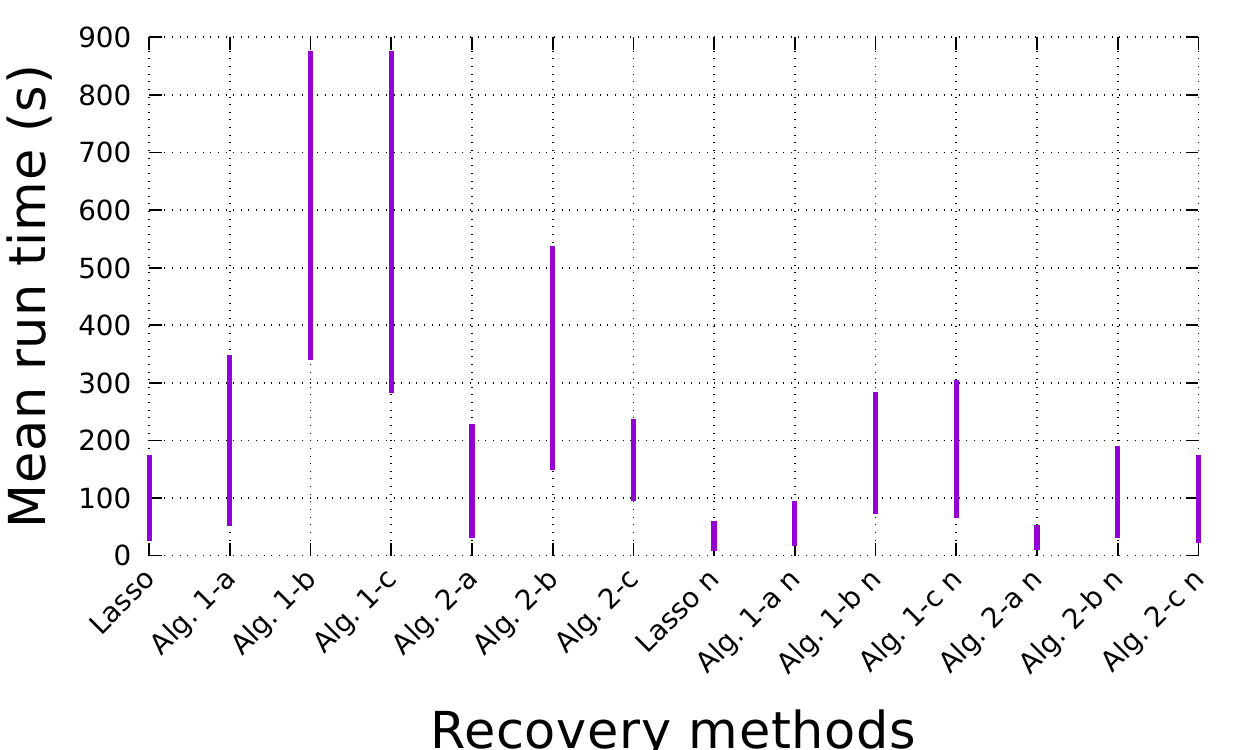}}
\caption{Lasso (via ADMM) vs Algorithms \ref{rwl1_fox} and \ref{rwl1_foxist}; $n=256$, $m=100$.}\label{fig:2}
\end{figure}
The first step of Algorithm \ref{rwl1_fox} requires to solve a Lasso problem, which can be  performed via iterative methods like IST and ADMM. This however entails a nested loop structure, which is generally not desired. We then propose a variant, summarized in Algorithm \ref{rwl1_foxist}, in which the solution of Lasso is substituted by a single IST step. $\soft_{\lambda w}$ indicates the soft thresholding operator \cite{for10} with componentwise threshold $\lambda w_i$.

\begin{algorithm}
\setstretch{1.3}
  \caption{Lasso IRL1 IST}\label{rwl1_foxist}
  \begin{algorithmic}[1] 
  \STATE  Initialize: $x(0)\in\R^n$, $\lambda>0$ $\tau>0$
  \FORALL{$t=1,\dots,T_{stop}$}
  \STATE $w_i(t)= g'(|x_i(t)|)$ for any $i\in \{1,\dots, n\}$
  \STATE $x(t+1)= \soft_{\lambda w(t)}[x(t)+\tau A^T(y-Ax(t))]$
    \ENDFOR     
  \end{algorithmic}
\end{algorithm} 

\begin{proposition}
Assume that $\tau \|A\|_2^2<1$ in Algorithm \ref{rwl1_foxist}. Given a sequence $(x(t),w(t))$ generated by Algorithm \ref{rwl1_foxist}, then $\fun(x(t),w(t))$ converges to a partial optimum.
\end{proposition}
\proof Algorithm \ref{rwl1_foxist} corresponds to the alternating minimization  of the tri-convex surrogate functional $\gun(x,w,b)=\fun(x,w)+\frac{1}{2}\left\|x-b \right\|_2^2-\frac{\tau}{2}\left\|A(x-b)\right\|_2^2$ adding the step $b(t)=x(t)$ (see \cite{for10}, for a review on the surrogate functional idea). For any block sub-problem, the minimum is unique under conditions discussed above, therefore $\gun(x(t),w(t),b(t))$, and thus $\fun(x(t),w(t))$, is strictly decreasing outside the solution set. This implies the thesis. \qed

In \cite{xu13,xu17}, more insight on alternating minimization/block coordinate descent of multi-convex or non-convex, non-differentiable problems can be found.
\subsection{Numerical experiment}
We finally show a numerical experiment to practically verify the efficiency of our algorithms.\footnote{The code to reproduce this experiment is available on https://github.com/sophie27/Lasso-l1-reweigthing.} We consider the experiment proposed in \cite[Section 3.1]{can08rew}, whose setting is as follows: $n=256$, $m=100$, $k\in[15,55]$; the unknown sparse signal has support generated uniformly at random, while its non-zero positions are randomly generated from a standard Gaussian distribution $\mathcal{N}(0,1)$;  the entries of the sensing matrix $A$ are randomly generated from a Gaussian distribution $\mathcal{N}(0,\frac{1}{m})$. We consider the penalties $g$'s defined in a), b), and c) in Section \ref{sub:bic} with $\varepsilon=10^{-1}$, $q=1/2, \alpha=2$.

We use ADMM to solve the classical Lasso and the Step 3 of Algorithm \ref{rwl1_fox}. We set $x(0)=(0,\dots,0)^T$, $\tau=2.5\times 10^{-1}$. We consider the noise-free case and the case when the signal-to-noise ratio is SNR=25dB; $\lambda$ is respectively set to $10^{-5}$ and $10^{-4}$. The algorithms are stopped whenever $\|x(t)-x(t-1)\|_2<10^{-5}$.  At most two reweighting iterations are performed for Algorithm \ref{rwl1_fox}. As noticed in \cite{can08rew}, in fact, the most of the enhancement is obtained within the first few reweighting iterations. The shown results are averaged over 100 runs.

In Figures \ref{fig:1}- \ref{fig:2} we show the performance of the proposed algorithms compared to Lasso, varying the sparsity level. Let $\xtrue$ and $\widehat{x}$ be the original signal and the estimation, respectively. The performance metrics are: the empirical probability of recovery for the noise-free case  (as in \cite{can08rew}, we define recovery the case $\left\|\xtrue-\widehat{x}\right\|_{\infty}<10^{-3}$); the relative square error $\frac{\left\|\xtrue-\widehat{x}\right\|_2^2}{\left\|\xtrue\right\|_2^2}$; the total number of iterations and run time. Simulations are performed on an AMD Opteron 6276, 2.30 GHz Processor. 

In Figures \ref{fig:1}-\ref{fig:2}, we can appreciate a general improvement obtained by Algorithms \ref{rwl1_fox} and \ref{rwl1_foxist}  with respect to Lasso in terms of recovery and relative square error, in particular using the log penalty a). In the noise-free case, the smaller $\lambda$ generates a more precise solution, but yields to slower convergence. In the noisy case, the larger $\lambda$ helps to tolerate noise and at the same time accelerates the procedure. Figure \ref{fig:2}(a) shows the ranges (in seconds) of the mean run times for each considered $k$. This is consistent with the number of iterations (Figure \ref{fig:1}(c)- Figure \ref{fig:2}(b)).

Clearly, Lasso requires less iterations than Algorithm \ref{rwl1_fox}, which is more accurate at the price of a longer run time. Algorithm \ref{rwl1_foxist} reduces the run time without substantial loss of accuracy. Algorithm \ref{rwl1_foxist}-a is the best choice for the proposed experiment: it has the best recovery accuracy in the noise free case (see Figure \ref{fig:1}(a)) and run time very close to Lasso; it only loses some accuracy for large $k$ in the noisy setting.
\section{Conclusions}
In this letter, we have proved that a Lasso iterative $\ell_1$ reweighting algorithm corresponds to the alternating minimization of a biconvex functional. This  allows us the prove the numerical convergence of the iterates of the algorithm, that is, the distance between two successive iterates goes to zero. This is a stronger convergence with respect to previous results on iterative $\ell_1$ reweighting methods.  Moreover, we have proposed an IST-based alternative algorithm, which converges to a partial optimum and is practically very fast.

\bibliographystyle{plain}
\bibliography{refs}

\end{document}